\documentclass[]{article}
\usepackage[letterpaper]{geometry}
\usepackage{amta2016}
\usepackage{times}
\usepackage{latexsym}
\usepackage{amsmath, amssymb}
\usepackage{graphicx}
\usepackage{multirow}
\usepackage{xspace}

\usepackage{amsmath}
\usepackage{stackrel}
\usepackage{graphics}
\usepackage{array}
\usepackage[usenames,dvipsnames,svgnames,table]{xcolor}
\usepackage{ucs}
\usepackage[utf8]{inputenc}
\usepackage{color}
\usepackage{amssymb,amsmath,epsfig}
\usepackage{algorithm}

\graphicspath{ {fig/}{fig-chen/}{} }

\newcommand\BLEU{\textsc{Bleu}\xspace}
\newcommand\TER{\textsc{Ter}\xspace}

\title{Fast Domain Adaptation for Neural Machine Translation}

\begin{document}

\title{Fast Domain Adaptation for Neural Machine Translation}

\author{Markus Freitag \and Yaser Al-Onaizan\\
      {\tt IBM T.J. Watson Research Center\\
            1101 Kitchawan Rd, Yorktown Heights, NY 10598\\
                \{freitagm,onaizan\}@us.ibm.com}}

\maketitle
\pagestyle{empty}

\begin{abstract}
Neural Machine Translation (NMT) is a new approach for automatic translation of text from one human language into another.
The basic concept in NMT is to train a large Neural Network that maximizes the translation performance on a given parallel corpus.
NMT is gaining popularity in the research community because it outperformed traditional SMT approaches in several
translation tasks at WMT and other evaluation tasks/benchmarks at least for some language pairs. However, many of the enhancements in SMT over the years have not been incorporated into the NMT framework. 
In this paper, we focus on one such enhancement namely domain adaptation. We propose an approach for adapting a NMT system to a new domain.
The main idea behind domain adaptation is that the availability of large out-of-domain training data and a small in-domain training data.
We report significant gains with our proposed method in both automatic metrics and a human subjective evaluation metric on two language pairs.
With our adaptation method, we show large improvement on the new domain while the performance of our general domain only degrades slightly.
In addition, our approach is fast enough to adapt an already trained system to a new domain within few hours without the need to
retrain the NMT model on the combined data which usually takes several days/weeks depending on the volume of the data.
\end{abstract}

\section{Introduction}
\label{sec:introduction}
Due to the fact that Neural Machine Translation (NMT) is reaching comparable or even better performance compared to the traditional statistical machine translation (SMT) models \cite{jean+:2015,luong+:2015}, it has become very popular in the recent years~\cite{kalchbrenner+blunsom:2013,sutskever+:2014,bahdanau+:2014}. 
With the great success of NMT, new challenges arise which have already been address with reasonable success in traditional SMT. One of the challenges is domain adaptation.
In a typical domain adaptation setup such as ours, we have a large amount of out-of-domain bilingual training data for which we already have a trained neural network model (baseline). 
Given only an additional small amount of in-domain data, the challenge is to improve the translation performance on the new domain without deteriorating the performance on the general domain significantly. One approach one might take is to combine the in-domain data with the out-of-domain data and train the NMT model from scratch. However, there are two main problems with that approach. First, 
training a neural machine translation system on large data sets can take several weeks and training a new model based on the combined training data is time consuming.
Second, since the in-domain data is relatively small, the out-of-domain data will tend to dominate the training data and hence the learned model will not perform as well on the in-domain test data.
In this paper, we reuse the already trained out-of-domain system and continue training only on the small portion of in-domain data similar to \cite{luong2015stanford}. While doing this, we
adapt the parameters of the neural network model to the new domain. Instead of relying completely on the adapted (further-trained) model and over fitting on
the in-domain data, we decode using an ensemble of the baseline model and the adapted model which tends to perform well on the in-domain data without
deteriorating the performance on the baseline general domain.

\section{Related Work}
\label{sec:related_work}
Domain adaptation has been an active research topic for the traditional SMT approach in the last few years.
The existing domain adaptation methods can be roughly divided into three different categories.

First, the out-of-domain training data can be scored by a model built only on the in-domain training data. Based on the scores, we can either use
a certain amount of best scoring out-of-domain training data to build a new translation system or assign a weight to each sentence which determines its contribution towards the  training a new system. In SMT, this has been done for language model training \cite{gao2002toward,moore2010intelligent} and translation model
training~\cite{matsoukas2009discriminative,foster2010discriminative,axelrod2011domain,mansour14:weighting}. In contrast to SMT, training a NMT system from scratch is time consuming and can easily take several weeks.

Second, different methods of interpolating in-domain and out-of-domain models \cite{lu2007improving,koehn2007experiments,foster2007mixture} have been proposed.
A widely used approach is to train an additional SMT system based only on the in-domain data in addition to the existing out-of-domain SMT system. 
By interpolating the phrase tables, the in-domain data can be integrated into the general system. In NMT, we do not have any phrase tables and can not use
this method. Nevertheless, integrating the in-domain data with interpolation is faster than building a system from scratch. 

The third approach is called semi-supervised training, where  a large in-domain monolingual data is first translated with a machine translation engine into a different language to generate  parallel data.
The automatic translations have been used for retraining the language model and/or the translation model \cite{ueffing2007transductive,schwenk2008investigations,huck2011lightly}.
Parallel data can be created also by back-translating monolingual target language into the source language creating additional parallel data\cite{sennrich2015improving}. The additional parallel training data can be used to train the NMT and obtain. \cite{sennrich2015improving} report 
 substantial improvements when a large amount of back-translated parallel data is used. However, as we mentioned before retraining the NMT model with large training data takes time and in this case it is even more time consuming since we first need to back-translate the target monolingual data and then build a system based on the combination of both the original parallel data and the back-translated data.

For neural machine translation, \cite{luong2015stanford} proposed to adapt an already existing NMT system to a new domain with
further training on the in-domain data only. The authors report an absolute gain of 3.8 \BLEU points compared to using an original model without further training.
In our work, we utilize the same approach but ensemble the further trained model with the original model. In addition, we report results on the out-of-domain
test sets and show how degradation of the translation performance on the out-of-domain can be avoided. We further show how to avoid over-fitting on the
in-domain training data and analyze how many additional epochs are needed to adapt the model to the new domain. We compare our adapted models with
models either trained on the combined training data or the in-domain training data only and report results for different amount of in-domain training data.

\section{Neural Machine Translation}
\label{sec:nmt}
In all our experiments, we use our in-house attention-based NMT implementation which is similar to ~\cite{bahdanau+:2014,mi2016vocabulary} 
The approach is based on an encoder-decoder network.
The encoder employs a bi-directional RNN to 
encode the source sentence ${\bf{x}}=({x_1, ... , x_l})$ 
into a sequence of hidden states ${\bf{h}}=({h_1, ..., h_l})$, where $l$ is the length of the source sentence.
Each $h_i$ is a concatenation of a left-to-right $\overrightarrow{h_i}$
and a right-to-left $\overleftarrow{h_i}$ RNN:
\[
h_{i} = 
\begin{bmatrix}
\overleftarrow{h}_i \\ 
\overrightarrow{h}_i \\
\end{bmatrix}
=
\begin{bmatrix}
\overleftarrow{f}(x_i, \overleftarrow{h}_{i+1}) \\
\overrightarrow{f}(x_i, \overrightarrow{h}_{i-1}) \\
\end{bmatrix}
\]
where $\overleftarrow{f}$ and $\overrightarrow{f}$ 
are two gated recurrent units (GRU) proposed by~\cite{cho+:2014_gru}.

Given the encoded ${\bf h}$, the decoder predicts the target translation
by maximizing the conditional log-probability of the 
correct translation ${\bf y^*} = (y^*_1, ... y^*_m)$, where 
$m$ is the length of the target. At each time $t$, 
the probability of each word $y_t$ from a target vocabulary $V_y$ is:
\begin{equation}
\label{eq:py}
p(y_t|{\bf h}, y^*_{t-1}..y^*_1) = g(s_t, y^*_{t-1}, H_{t}),
\end{equation}
where $g$ is 
a two layer feed-forward neural network 
over the embedding of the previous target word $y^*_{t-1}$,  
the hidden state $s_t$, and the weighted sum of ${\bf h}$ ($H_{t}$). 

Before we compute $s_t$ and $H_t$, we first covert $s_{t-1}$ and 
the embedding of $y^*_{t-1}$ into an intermediate state $s'_t$ with a GRU $u$ as:
\begin{equation}
s'_t = u(s_{t-1}, y^*_{t-1}).
\end{equation}
Then we have $s_t$ as:
\begin{equation}
s_t = q(s'_{t}, H_{t})
\end{equation}
where $q$ is a GRU.
And the $H_{t}$ is computed as:
\begin{equation}
H_t = 
\begin{bmatrix}
\sum_{i=1}^{l}{(\alpha_{t,i} \cdot \overleftarrow{h}_i)} \\
\sum_{i=1}^{l}{(\alpha_{t,i} \cdot \overrightarrow{h}_i)} \\
\end{bmatrix},
\end{equation}
The alignment weights, $\alpha$ in $H_t$, are computed with a two layer feed-forward neural network $r$:
\begin{equation}
\alpha_{t,i} = \frac{\exp\{r(s'_{t}, h_{i})\}}{\sum_{j=1}^{l}{\exp\{r(s'_{t}, h_{j})\}}}
\end{equation}

\section{Domain Adaptation}
\label{sec:domain_adaptation}
Our objectives in domain adaptation are two fold: (1) build an adapted system quickly (2) build a system that performs
well on the in-domain test data without significantly degrading the system on a general domain.  
One possible approach to domain adaptation is to mix (possibly with a higher weight) the in-domain with the large out-of-domain data 
and retrain the system from scratch. However, training a NMT system on large amounts of parallel data (typically $>$4 million sentence pairs)
can take several weeks. Therefore, we propose a method that doesn't require retraining on the large out-of-domain data which
we can do relatively quickly. Hence, achieving our two objectives.

Our approach re-uses the already trained \textit{baseline} model and continues the training for several additional epochs but only on the small amount
of in-domain training data. We call this kind of further training a \textit{continue} model. Depending on the amount
of in-domain training data, the continue model can over-fit on the new training data. In general over-fitting means that the model performs
excellent on the training data, but worse on any other unseen data. To overcome this
problem, we ensemble the continue model with the baseline model. This has the positive side effect that we do not only get
better translations for the new domain, but also stay close to the baseline model which performs well in general. As the amount of
in-domain training data is usually small, we can quickly adapt our baseline model to a different domain.

\section{Experiments}
\label{sec:experiments}
In all our experiments, we use the NMT approach as described in Section~\ref{sec:nmt}.
We limit our source and target vocabularies to
be the top $N$ most frequent words for each side accordingly. Words not in these vocabularies are mapped into a special unknown token UNK. 
During translation, we write the alignments (from the attention mechanism) and use these to replace the unknown tokens either with
potential targets (obtained from an IBM model 1 dictionary trained on the parallel data or from the SMT phrase table) or with the source word itself or a transliteration of it (if no target was found in the dictionary, i.e., the word is a genuine OOV). 
We use an embedding dimension of 620 and fix the RNN GRU layers to be of 1000 cells each. For
the training procedure, we use SGD~\cite{bishop1995neural} to update the model parameters with a mini-batch size of 64.
The training data is shuffled after each epoch.
All experiments are evaluated with both \BLEU \cite{papineni2002bleu} and \TER \cite{snover2006study} (both are case-sensitive).

\subsection{German$\rightarrow$English}

For the German$\rightarrow$English translation task, we use an 
already trained out-of-domain NMT system (vocabulary size $N$=100K) trained on the WMT 2015 training data \cite{bojar-EtAl:2015:WMT} (3.9M parallel sentences).
As in-domain training data, we use the TED talks from the IWSLT 2015 evaluation campaign \cite{cettoloiwslt} (194K parallel sentences).
Corpus statistics can be found in Table~\ref{tab:corpora:Stats-GeEn}.
The data is tokenized and the German text is preprocessed by splitting German compound words with the frequency-based method as described in \cite{koehn2003empirical}.
We use our in-house language identification tool to remove sentence pairs where either the source or the target is assigned the wrong language by our language ID.

\begin{table}[ht!]
\begin{center}
\begin{tabular}{|l|l|cc|}
    \hline
    &  &  German    &   English      \\ \hline \hline
  &Sentences    & \multicolumn{2}{c|}{3.9M}\\ 
   WMT 2015 & Running Words& 89M & 93M \\ 
      & Vocabulary   & 895K & 698K \\ \hline 
   &Sentences    & \multicolumn{2}{c|}{194k}\\ 
    IWSLT 2015  & Running Words& 3.7M & 3.9M \\ 
     & Vocabulary   & 102K & 61K \\ \hline
\end{tabular}
\caption{German$\rightarrow$English corpus statistics for in-domain (IWSLT 2015) and out-of-domain (WMT 2015) parallel training data.}
\label{tab:corpora:Stats-GeEn}
\end{center}
\end{table}

Experimental results can be found in Table~\ref{tab:results}. The translation quality of a NMT system trained only on the in-domain data 
is not satisfying. In fact, it performs even worse on both
test sets compared to the baseline model which is only trained on the out-of-domain data. By continuing the training of the baseline model on the in-domain data only, we get a
gain of 4.4 points in \BLEU and 3.1 points in \TER on the in-domain test set tst2013 after the second epoch. Nevertheless, we lose
2.1 points in \BLEU and 3.9 points in \TER on the out-of-domain test set newstest2014. After continuing the epoch for 20 epochs, the
model tends to overfit and the performance of both test sets degrades.

\begin{table}[ht!]
\begin{center}
    \begin{tabular}{|l|c|cccc|}
        \hline
        \bf{system}& \bf{epoch} & \multicolumn{2}{c}{\bf{newstest2014}} & \multicolumn{2}{c|}{\bf{tst2013}} \\
                   & & \BLEU & \TER  & \BLEU & \TER \\ \hline \hline
        baseline (only out-of-domain)  & & 24.8 & 56.0 & 29.2 & 49.4\\ \hline
        only in-domain & & 13.2 & 74.3 & 28.4 & 52.2 \\        
        combined data & & 25.0 & 55.7 & 32.4 & 46.2 \\ \hline        
                 & 1 & 23.0 & 57.6 & 31.8 & 46.6 \\
        continue & 2 & 22.7 & 59.9 & \bf{33.6} & \bf{46.3} \\ 
                 & 20 & 20.3 & 63.0 & 30.5 & 49.1 \\ \hline
           & 1 & 25.4 & 54.9 & 32.2 & 46.3 \\
       ensemble of baseline + continue & 2 & 25.6 & 55.6 & \bf{33.3} & \bf{45.3} \\
              & 20 & 24.6 & 56.6 & 33.4 & 45.7 \\ \hline
\end{tabular}
\caption{Adaptation results for the German$\rightarrow$English translation task: tst2013 is the in-domain test set and newstest2014 is the out-of-domain test set. The combined data is the concatenation of the in-domain and out-of-domain training data.}
\label{tab:results}
\end{center}
\end{table}

To avoid over fitting and to keep the out-of-domain translation quality close to the baseline, we ensemble the continue model with the baseline model. After 20 epochs,
we only lose 0.2 points in \BLEU and 0.6 points in \TER on the out-of-domain test set while we gain 4.2 points in \BLEU and
3.7 points in \TER on tst2013.
Each epoch of the continue training takes 1.8 hours. In fact, with only two epochs, we already have a very good performing system on the in-domain data.
At the same time, the loss of translation quality on the out-of-domain test set is minimal (i.e., negligible).
In fact, we get a gain of 0.7 points in \BLEU while losing 0.6 points in \TER on our out-of-domain test set.

\begin{figure}[ht!]
\centering
\resizebox{\linewidth}{!}{\input{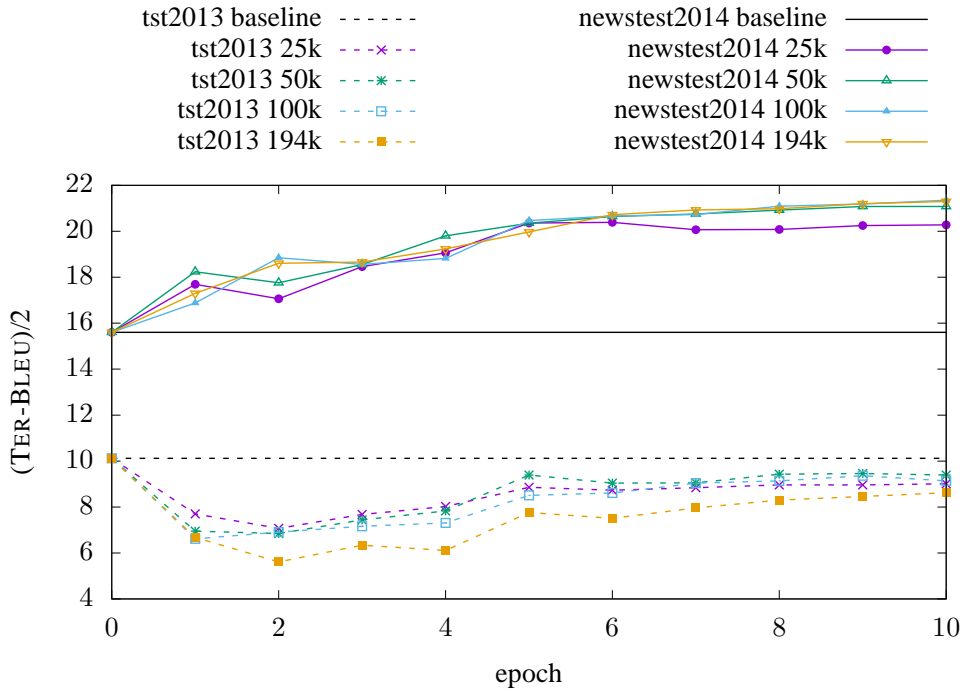}}
\vspace{-2em}
\caption{German$\rightarrow$English: Learning curve of the continue training. Scores are given in (\TER-\BLEU)/2 (lower is better). tst2013 is our in-domain and newstest2014 is our out-of-domain test set.
The baseline model is only trained on the large amount of out-of-domain data.}
\label{fig:continue}
\end{figure}

Figure~\ref{fig:continue} illustrates the learning curve of the continue training for different sizes of in-domain training data.
For all setups, the translation quality massively drops on the out-of-domain test set. Further, the performance of the in-domain test set
degrades as the neural network over-fits on the in-domain training data already after epoch 2. 

\begin{figure}[ht!]
\centering
\resizebox{\linewidth}{!}{\input{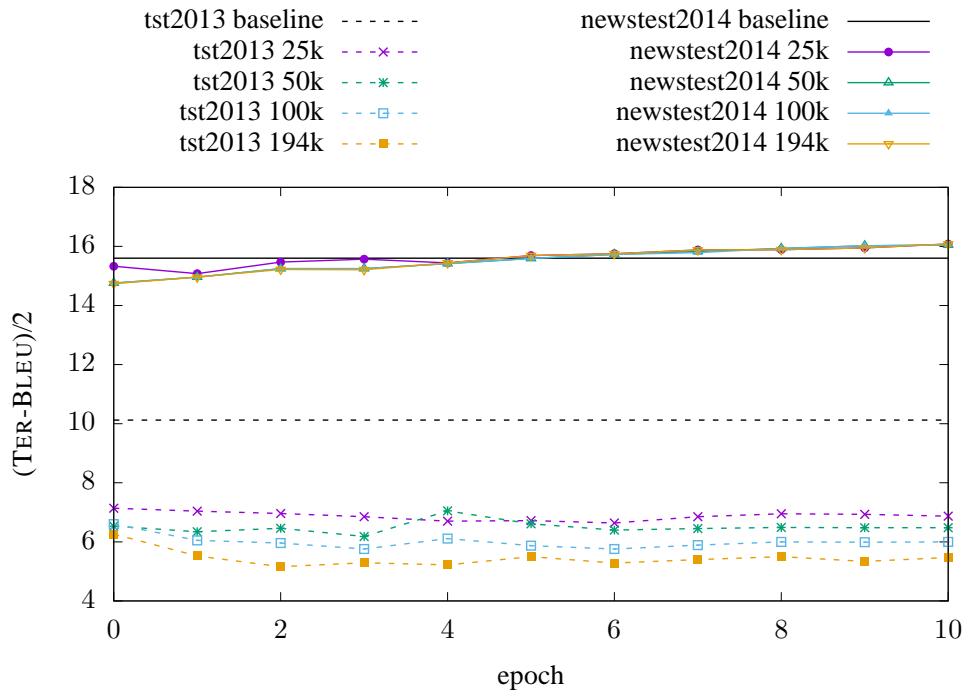}}
\vspace{-2em}
\caption{German$\rightarrow$English: Learning curve of the ensemble of 2 models: continue training (cf.~Figure~\ref{fig:continue}) and baseline model. tst2014 is the in-domain; newstest2014 the out-of-domain test set.}
\label{fig:continue_ensemble}
\end{figure}

To study the impact of the in-domain data size on the quality if the adapted model, we report results for different sizes of the in-domain data. Figure~\ref{fig:continue_ensemble} shows the learning curve of the ensemble of the baseline and the continue model for different sizes of in-domain training data. The used in-domain data is a randomly selected subset of the entire pool of the in-domain data available to us. We also report the result when all of the in-domain data in the pool is used.
As shown in Figure~\ref{fig:continue_ensemble} the translation quality of the out-of-domain test set only degrades slightly for all the different sizes of the in-domain data we tried. However, the performance on the in-domain data significantly
improves, reaching its peak just after the second epoch. We do not lose any translation quality on the in-domain test set by continuing the training for more epochs.
Adding more in-domain data improved the score on the in-domain test set without seeing any significant degradation on the out-of-domain test set.

In addition to evaluating on automatic metrics, we also performed a subjective human evaluation where a human annotator assigns a score based on the quality of the translation. The judgments are done by an experienced annotator (a native speaker of German and a fluent speaker of English). 
We ask our  annotator to judge the translation output of different systems on a randomly selected  in-domain sample of 50 sentences (maximum sentence length 50). Each source sentence is presented to the annotator with all 3 different translations (baseline/ continue/ ensemble). The translation are presented in a blind fashion (i.e., the annotator is not aware of which system is which) and shuffled in random order. The evaluation is presented to the annotator via a web-page interface with all these translation pairs randomly ordered to disperse the three translations of each source sentence. The annotator judges
each translation from 0 (very bad) to 5 (near perfect).
The human evaluation results can be found in Table~\ref{tab:results_deen_human}. Both the
continue as well as the ensemble of the baseline with the continue model significantly outperforms the baseline model on the in-domain data. Contrary to the automatic scores,
the ensemble performs better compared to the continue model.

\begin{table}[ht!]
\begin{center}
    {\setlength{\tabcolsep}{.3em}
    \begin{tabular}{|l|r|r|r|r|r|r|r|}
        \hline
                 & 0 & 1 & 2  & 3  & 4  & 5  & Avg \\ \hline \hline
        baseline & 1 & 0 & 10 & 9 & 13  & 17  & 3.68 \\
        continue epoch 2 & 0 & 0 & 7 & 9 & 14 & 20  & 3.94 \\
        ensemble epoch 2 & 0 & 0 & 8 & 6 & 12 & 24 & 4.04 \\ \hline
\end{tabular}
\caption{German$\rightarrow$English: Human evaluation on an in-domain sample of 50 sentences. The annotator assigns each sentence a score between 0-5 (higher is better).}
\label{tab:results_deen_human}
}
\end{center}
\end{table}

We compare the training times of our different setups in Table~\ref{tab:training-time-deen}. Based on the automatic scores, it is sufficient to further
train the baseline model for 2 epochs to adapt the model to the new domain. For the case of having only 25K parallel in-domain training data, it
takes 30 minutes to adapt the model. If we use all available 192K sentences, the total training time is 3 hours and 40 minutes. By using all training
data (both in-domain and out-of-domain together), we need 7 epochs which sum up to a training time of 15 days and 11 hours.  

\begin{table}[ht!]
\begin{center}
\begin{tabular}{|l|l|}
    \hline
     \bf{training data}   & \bf{running time} \\ \hline \hline
    25K sentences & 15 minutes \\
    50K sentences & 30 minutes \\
    100K sentences & 1 hour \\
    192K sentences & 1 hour 50 minutes \\
    4.1M sentences & 53 hours \\ \hline
\end{tabular}
\caption{German$\rightarrow$English training times per epoch. The setup including both in-domain and out-of-domain training data has 4.1M parallel sentences.}
\label{tab:training-time-deen}
\end{center}
\end{table}

\subsection{Chinese$\rightarrow$English}

For the Chinese$\rightarrow$English experiments, we utilize a NMT system (vocabulary size $N$=500K) trained on 11.6 million out-of-domain sentences
from the DARPA BOLT project. We use 593k parallel sentences of internal in-domain data that is different to the BOLT informal news domain.
Corpus statistics can be found in Table~\ref{tab:corpora:Stats-ChEn}.

\begin{table}[ht!]
\begin{center}
\begin{tabular}{|l|l|cc|}
    \hline
    &  &  Chinese    &   English      \\ \hline \hline
  &Sentences    & \multicolumn{2}{c|}{11.6M}\\ 
   out-of-domain & Running Words& 302M & 330M \\ 
      & Vocabulary   & 545K & 536K \\ \hline 
   &Sentences    & \multicolumn{2}{c|}{593k}\\ 
    in-domain   & Running Words& 10.1M & 12.7M \\ 
     & Vocabulary   & 74K & 63K \\ \hline
\end{tabular}
\caption{Chinese$\rightarrow$English corpus statistics for in-domain and out-of-domain parallel training data.}
\label{tab:corpora:Stats-ChEn}
\end{center}
\end{table}

Experimental results can be found in Table~\ref{tab:results_chen}. Because the in-domain data is relatively large in this case, training a NMT model from scratch only on the in-domain data 
gives us similar performance on the in-domain test set compared to the baseline model that is trained only on the out-of-domain data. However,
the performance on the out-of-domain test set is significantly worse. By continuing the training of the baseline model only on the
in-domain data, we get an improvement of 9.5 points in \BLEU and 12.2 points in \TER on the in-domain test set after 6 epochs.
Unfortunately, the performance significantly drops on the out-of-domain test set. After 20 epochs, the performance
on the in-domain data only further improves slightly while losing much more on the out-of-domain test set.

To avoid significant degradation to the translation quality on the out-of-domain test set, we ensemble the continue and the baseline models.
After 6 epochs, we get a gain of 7.2 points in \BLEU and 10 points in \TER on the in-domain
test set while losing only slightly on the out-of-domain test set. After 20 epochs, the performance of the in-domain test set is similar
while losing additional 1.5 points in \BLEU and 1.1 points in \TER on the out-of-domain test set.

\begin{table}[t!]
\begin{center}
    \begin{tabular}{|l|c|cccc|}
        \hline
        \bf{system}& \bf{epoch} & \multicolumn{2}{c}{\bf{out-of-domain}} & \multicolumn{2}{c|}{\bf{in-domain}} \\
                   & & \BLEU & \TER  & \BLEU & \TER \\ \hline \hline
        baseline (only out-of-domain) & & 33.4 & 58.6 & 15.0 & 73.9\\ \hline
        only in-domain & & 17.6 & 72.0 & 15.2 & 73.5 \\
        combined data & & 32.6 & 57.5 & 20.5 & 65.0 \\ \hline
                 & 1 & 29.7 & 59.1 & 18.8 & 66.6 \\
        continue & 6 & 27.6 & 61.1 & \bf{24.5} & \bf{61.7} \\ 
                 & 20 & 25.0 & 63.1 & 24.9 & 61.7 \\ \hline
                & 1 & 33.2 & 56.4 & 17.8 & 68.1 \\
        ensemble of baseline + continue & 6 & 32.6 & 57.1 & \bf{22.2} & \bf{63.9} \\
              & 20 & 31.1 & 58.2 & 22.2 & 64.1 \\ \hline
\end{tabular}
\caption{Chinese$\rightarrow$English adaptation results. The adaptation has been utilized on 593k in-domain parallel sentences.}
\label{tab:results_chen}
\end{center}
\end{table}

\begin{figure}[ht!]
\centering
\resizebox{\linewidth}{!}{\input{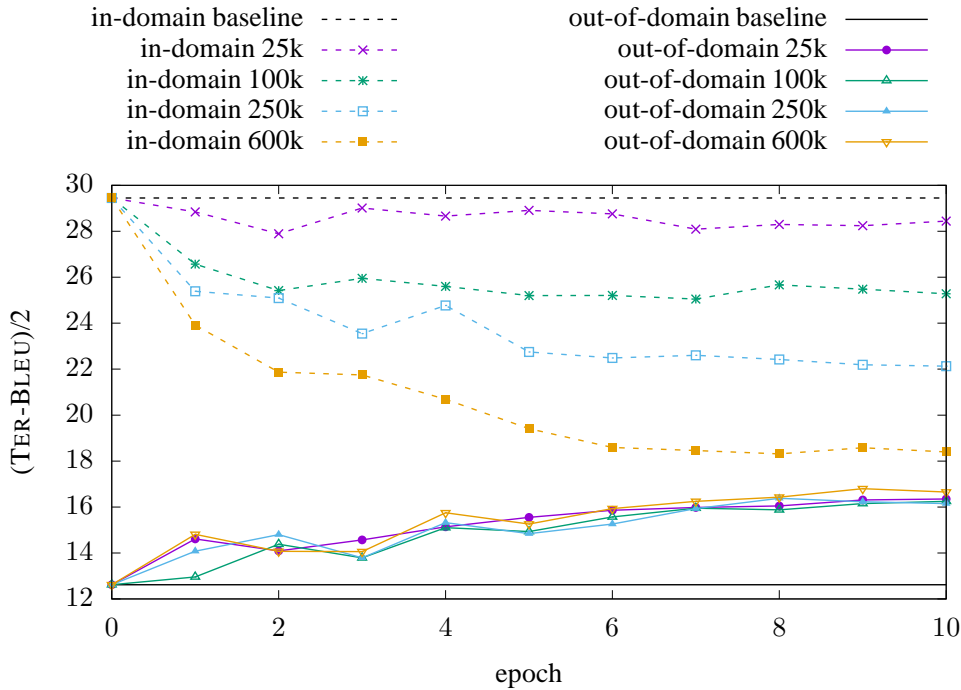}}
\vspace{-2em}
\caption{Chinese$\rightarrow$English: Learning curve of the continue training. Scores are given in (\TER-\BLEU)/2 (lower is better).
    The baseline is only trained on the large amount of out-of-domain training data.}
\label{fig:continue_chen}
\end{figure}

Figure~\ref{fig:continue_chen} illustrates the learning curves of the continue training for different sizes of in-domain training data.
Adding more parallel in-domain training data helps to improve the performance on the in-domain test set. 
For all different training sizes, the translation quality drops similar on the out-of-domain test set.  

\begin{figure}[ht!]
\centering
\resizebox{\linewidth}{!}{\input{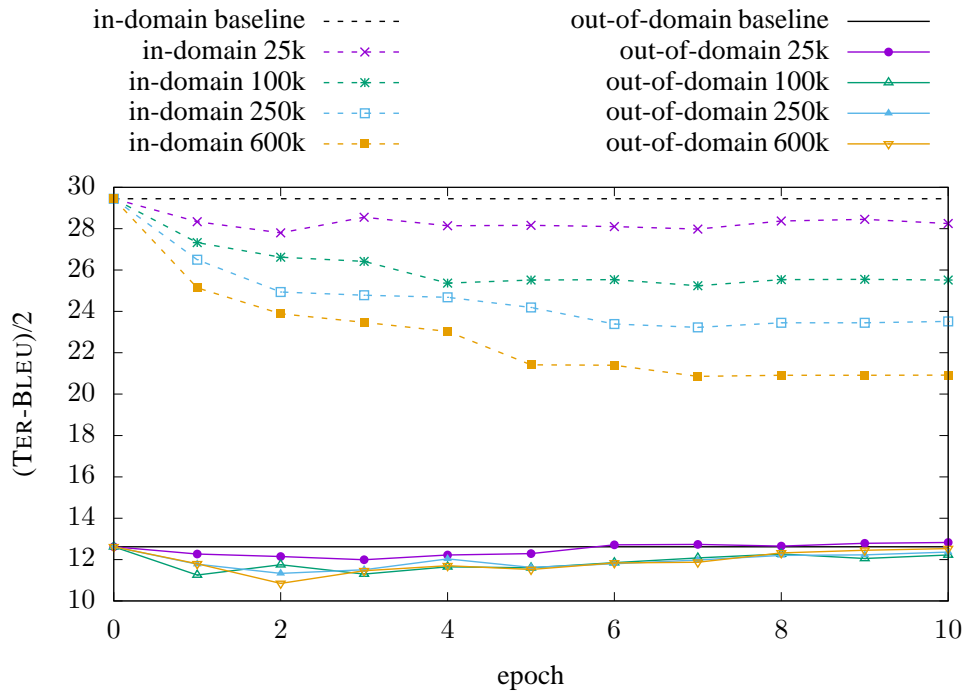}}
\vspace{-2em}
\caption{Chinese$\rightarrow$English: Learning curve of the ensemble of 2 models: the continue training (cf.~Figure~\ref{fig:continue_chen}) with the baseline model. The smaller training sets are
random subsets of the complete in-domain training data.}
\label{fig:continue_ensemble_chen}
\end{figure}

Figure~\ref{fig:continue_ensemble_chen} shows the learning curves of the ensemble of the baseline and the continue model for different sizes of in-domain training data.
For all training sizes, the translation quality of the out-of-domain test set only degrades slightly. Nevertheless, the performance on the in-domain data significantly
improves. We reach a saturation by continuing the training for several epochs on both test sets. Adding more in-domain data improves the score on the in-domain test set.

Human judgment was performed (cf.~Table~\ref{tab:results_chen_human}) by another experienced annotator (Chinese native speaker whose also fluent in English)
on a randomly selected sample of 50 in-domain sentences. As in the German$\rightarrow$English case, the annotator assigns a (0-5) score to each translation.
Both, the continue as well as the ensemble of the baseline with the continue model outperforms the baseline model.
Furthermore, the ensemble of the continue model with the baseline model outperforms the continue training on its own. 

\begin{table}[ht!]
\begin{center}
    {\setlength{\tabcolsep}{.3em}
    \begin{tabular}{|l|r|r|r|r|r|r|r|}
        \hline
                 & 0 & 1 & 2  & 3  & 4  & 5  & Avg \\ \hline \hline
        baseline & 0 & 0 & 24 & 14 & 9  & 3  & 2.82 \\
        continue epoch 6 & 0 & 0 & 19 & 11 & 11 & 9  & 3.20 \\ 
        ensemble epoch 6 & 0 & 0 & 17 & 10 & 11 & 12 & 3.36 \\ \hline
\end{tabular}
\caption{Human evaluation on a 50 sentence Chinese$\rightarrow$English in-domain sample. The annotator assigns each sentence a score between 0-5 (higher is better).}
\label{tab:results_chen_human}
}
\end{center}
\end{table}

A comparison of the training times of our different setups can be found in Table~\ref{tab:training-time-chen}. Based on our experiments, it is sufficient to further train
the baseline for 6 epochs to adapt the neural net to our new domain. By using all available in-domain training data, we have a total training time of 23 hours. The training time
for a system based on both in-domain and out-of-domain training data needs already 77 hours and 30 min for one epoch. We trained the combined system for 8 epochs which sum up
to a total training time of 620 hours (25 days and 20 hours).

\begin{table}[ht!]
\begin{center}
\begin{tabular}{|l|l|}
    \hline
     \bf{training data}   & \bf{running time} \\ \hline \hline
    25K sentences & 10 minutes \\
    100K sentences & 40 minutes \\
    250K sentences & 1 hour 30 minutes \\
    600K sentences & 3 hour 50 minutes \\
    12.2M sentences & 77 hours 20 minutes \\ \hline
\end{tabular}
\caption{Chinese$\rightarrow$English training times per epoch. The setup including both in-domain and out-of-domain training data has 12.2M parallel sentences.}
\label{tab:training-time-chen}
\end{center}
\end{table}

\section{Conclusion}
\label{sec:conclusion}
We presented an approach for a fast and efficient way to adapt an already existing NMT system to a new domain without degradation of the translation quality on the out-of-domain test set.
Our proposed method is based on two main steps: (a) train a  model on only on the in-domain data, but initializing all parameters of the the neural network model
with the one from the existing baseline model that is trained on the large amount of out-of-domain training data (in other words, we continue the training of the baseline model only on the in-domain data); (b) ensemble the continue model with the baseline model at decoding time. While step (a) can lead to significant gains of up to 9.9 points in \BLEU and 12.2 points in \TER on the in-domain test set. However, it comes at the expense of significant degradation to the the translation quality on the original out-of-domain test set. Furthermore, the continue model tends to overfit the small
amount of in-domain training data and even degrades translation quality on the in-domain test sets if you train beyond one or two epochs.

Step (b) (i.e., ensembling of the baseline model with the continue
model)  ensures that the performance does not drop significantly on the out-of-domain test set while still getting significant improvements of up to 7.2 points in \BLEU
and 10 points in \TER on the in-domain test set. Even after only few epochs of continue training, we get results that are close to the results obtained
after 20 epoch. We also show significant improvements on on human judgment as well. We presented results on two diverse language pairs German$\rightarrow$English and Chinese$\rightarrow$English (usually very challenging pairs for machine translation). 

\bibliographystyle{apalike}
\bibliography{amta2016}

\end{document}